\documentclass[10pt,twocolumn,letterpaper]{article}

\usepackage{cvpr}      
\usepackage[accsupp]{axessibility}
\usepackage{kotex} 
\usepackage{graphicx}
\usepackage{multirow}
\usepackage{array, tabularx, booktabs, makecell}
\usepackage{lipsum}
\definecolor{cvprblue}{rgb}{0.21,0.49,0.74}
\usepackage[pagebackref,breaklinks,colorlinks,allcolors=cvprblue]{hyperref}

\def\paperID{19287} 
\def\confName{CVPR}
\def\confYear{2026}

\newcommand\blfootnote[1]{
  \begingroup
  \renewcommand\thefootnote{}\footnote{#1}
  \addtocounter{footnote}{-1}
  \endgroup
}

\title{FEAT: Fashion Editing and Try-On from Any Design}

\author{
Soye Kwon$^1$\hspace{6mm}
Keonyoung Lee$^2$\hspace{6mm}
Dahuin Jung$^{3}$\thanks{Corresponding authors}\hspace{6mm}
Jaekoo Lee$^1$\footnotemark[1]\\
$^1$Department of Computer Science, Kookmin University \\
$^2$School of Software, Soongsil University \\
$^3$Department of Artificial Intelligence, Chung-Ang University \\ 
{\tt\small soye0710@kookmin.ac.kr, joseph12752@gmail.com, dahuinjung@cau.ac.kr, jaekoo@kookmin.ac.kr}
}

\begin{document}

\twocolumn[{%
\renewcommand\twocolumn[1][]{#1}%
\maketitle
\includegraphics[width=\linewidth]{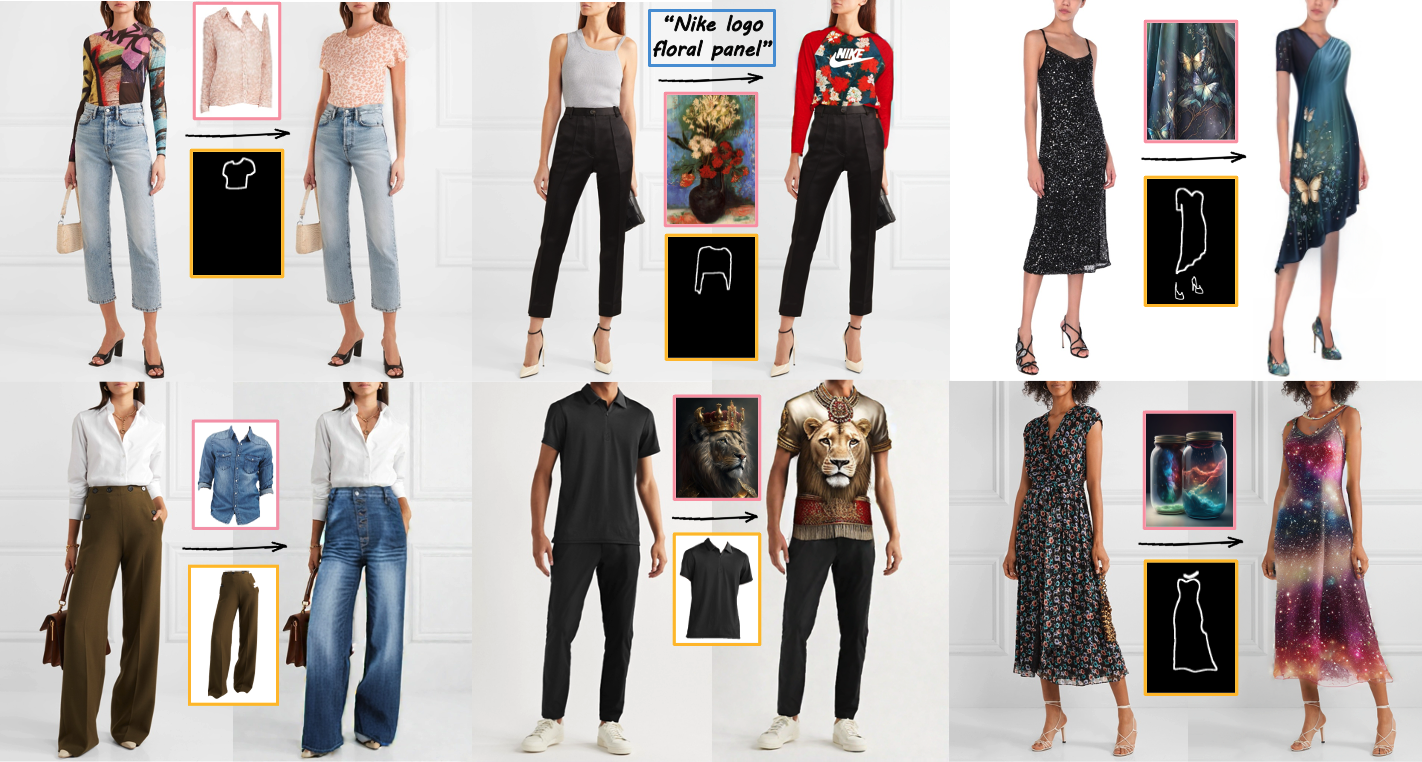}
\captionof{figure}{Examples of \textit{\textbf{FEAT} (\textbf{F}ashion \textbf{E}diting \textbf{A}nd \textbf{T}ry-On from Any Design)}. Yellow box: target prompt; Pink box: source prompt; Blue box: text prompt.
 \vspace{1em}}
\label{fig:teaser}
}]

\blfootnote{$^*$ Corresponding authors}


\begin{abstract}
Fashion design aims to express a designer’s creative intent and to depict how garments interact with the human body. Recent methods condition on multimodal inputs to support garment editing and virtual try-on. However, existing methods still (i) confine design to garment-related images, excluding creative design sources such as artwork, abstract imagery, and natural photographs, and (ii) cannot support complete outfits, including accessories. We present \textbf{FEAT} (\textbf{F}ashion \textbf{E}diting \textbf{A}nd \textbf{T}ry-On from Any Design), a method that enables editing and try-on across garments and accessories using diverse design sources. To achieve this, we introduce Disentangled Dual Injection (DDI). It takes both apparel and non-apparel design sources and selectively injects design cues via content and style disentanglement. Furthermore, we propose Orthogonal-Guided Noise Fusion (OGNF), a training-free mechanism that removes residual garments via orthogonal projection and applies region-specific noise strategies to enable virtual try-on for both garments and accessories. Extensive experiments demonstrate that \textit{FEAT} achieves state-of-the-art performance in design flexibility, prompt consistency, and visual realism.
\end{abstract}    
\section{Introduction}
\label{sec:intro}
Fashion is a domain where human creativity and aesthetic intent are rendered most directly, bridging artistic imagination and physical realization. In the digital era, AI-driven fashion design has progressed from merely synthesizing garment images to integrated pipelines that support virtual try-on (VTON), enabling realistic, user-controllable previews prior to production~\cite{multimodalgarment, fashiontex, picture}. 
Notably, recent approaches~\cite{picture, kim2025bayesian, fashiontex} accept multimodal inputs—visual exemplars and textual prompts—so users can articulate composite design intent and preview results, further connecting creative ideation to consumer experience.

Despite these advancements, current methods~\cite{fashiontex, picture} still exhibit critical limitations. (i) Most methods rely on garment-specific images, limiting creative design possibilities from broader sources such as artwork, natural imagery, or abstract concepts. (ii) Existing techniques also focus mainly on clothing, neglecting outfit compositions with accessories such as footwear, bags, and necklaces, which restricts practical applicability and holistic user experiences. To move beyond these limitations, we advocate a paradigm that embraces diverse design sources and compositional flexibility of fashion items.

Therefore, we present \textit{\textbf{FEAT} (\textbf{F}ashion \textbf{E}diting \textbf{A}nd \textbf{T}ry-On from Any Design)}, a diffusion-based method that enables editing and try-on across both garments and accessories using diverse design sources. In \textit{FEAT}, we distinguish between content and style, two primary attributes of fashion design. Content refers to the subject’s “what” (e.g., shape, contour, outline), whereas style denotes its “how” (e.g., colors and textures)~\cite{gatys2016image, kim2025improving, li2017laplacian}.

\begin{figure}[t]
    \centering
    \includegraphics[width=0.48\textwidth]{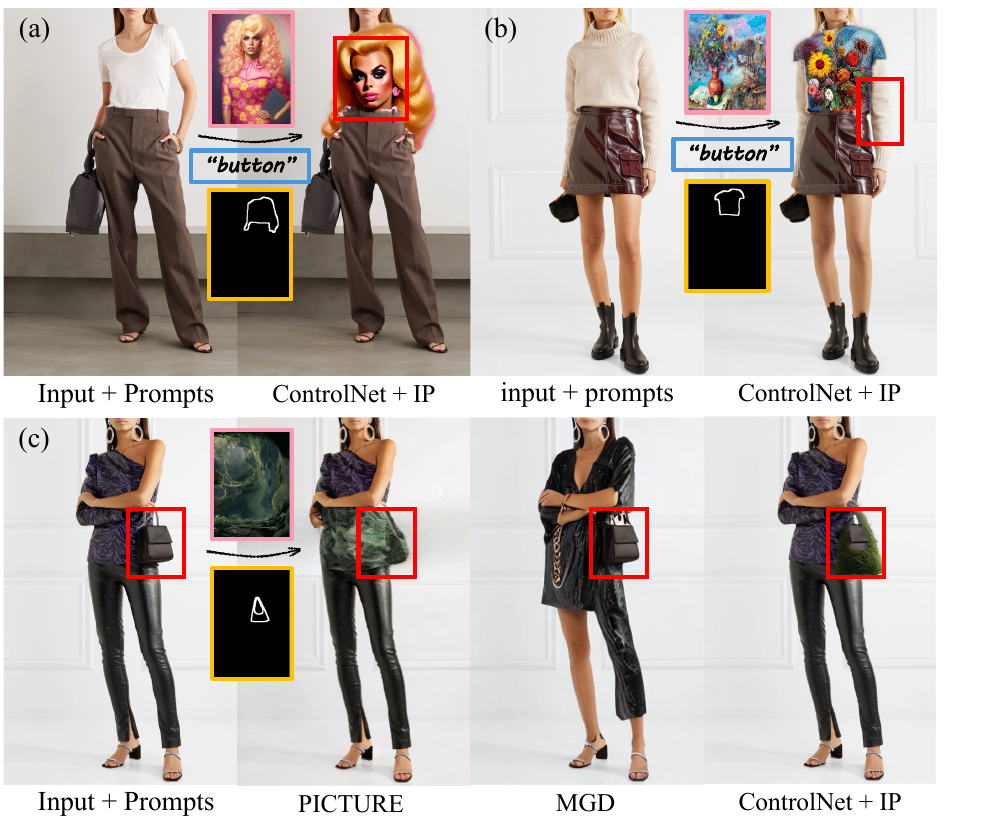}
    \caption{Problems and limitations of existing methods.}
    \label{fig:problem}
\end{figure}

As shown in Fig.~\ref{fig:problem}(a), directly injecting entangled content and style features from an image prompt through an IP-Adapter~\cite{ip-adapter} often over-amplifies content cues (e.g., faces), leading to content leakage~\cite{instantstyle,puff-net}. As a result, other conditioning signals, such as sketches and text prompts, can be suppressed.
Prior work~\cite{instantstyle} mitigates leakage by removing content information altogether, but this sacrifices practical utility because users often want the content elements from the image prompt to be reflected in the garment. To address this limitation, \textit{FEAT} introduces \textit{Disentangled Dual Injection (DDI)}, which separates content and style in the image prompt and injects them selectively. By routing style and content features to different U-Net blocks according to their roles, DDI mitigates content leakage while preserving structural cues and enabling user-controlled trade-offs through adjustable content and style scales.

Combining ControlNet~\cite{controlnet} with IP-Adapter~\cite{ip-adapter} often fails to fully replace clothing—an essential requirement for VTON—leaving residual garments (Fig.~\ref{fig:problem}(b)). Existing methods that rely on garment-specific datasets~\cite{multimodalgarment, picture, fashiontex} also suffer from limitations in scalability and practical applicability, particularly when handling unseen fashion items (Fig.~\ref{fig:problem}(c)). Moreover, collecting a comprehensive fashion-item dataset is prohibitively expensive. To overcome these challenges, we introduce a novel training-free mechanism intended for full-outfit try-on: \textit{Orthogonal-Guided Noise Fusion (OGNF)}. \textit{OGNF} removes the garment regions to be replaced via orthogonal projection~\cite{pearson1901closestfit} and applies region-specific noise strategies for try-on. Thus, our approach reduces reliance on extensive dataset curation and retraining, improving both practical usability and scalability.

User studies, while informative, often lack objectivity and reproducibility. To address this, we establish a new quantitative evaluation metric incorporating Chamfer Distance (CD)~\cite{ssim}, CLIP-based image–text similarity~\cite{clip}, and Elo ratings computed via a GPT-4V-based oracle model~\cite{gpteval3d}. This objective and reproducible metric enables consistent performance comparisons across diverse scenarios and demonstrates that our method outperforms recent state-of-the-art approaches~\cite{controlnet, picture, multimodalgarment}. To support reproducibility and future work, we publicly release all code, generated results, and a new fashion editing and try-on dataset containing diverse items generated with \textit{FEAT}.

\begin{itemize}
\item We propose \textit{DDI}, which enables fine-grained integration of diverse design sources, facilitating broader and more controllable editing.
\item We introduce \textit{OGNF}, a training-free approach that supports full-outfit try-on beyond garments, including accessories and multi-item compositions.
\item We publicly release a virtual fitting dataset covering diverse fashion items, providing a critical resource for systematic benchmarking and advancement in VTON.
\item We establish a rigorous evaluation using CD, CLIP image/text scores, user studies, and oracle-based metrics to benchmark VTON methods reliably.

\end{itemize}

\section{Related Work}
\noindent {\bf Multimodal Conditioning for Diffusion Models.}
Recent diffusion models have begun exploring multimodal conditioning to overcome the difficulty of achieving fine-grained control with text prompts alone~\cite{controlnet, uni-controlnet, ip-adapter, instantstyle}.
ControlNet~\cite{controlnet} improves controllability using an additional spatial branch, but requires a separate branch for each modality. Uni-ControlNet~\cite{uni-controlnet} mitigates this by fine-tuning two generic adapters. However, these methods~\cite{controlnet, 3d_controllable, uni-controlnet} incorporate image prompt only partially, leading to limited visual faithfulness.
To address this, IP-Adapter~\cite{ip-adapter} introduces a decoupled cross-attention mechanism to enhance the expressiveness of image prompts, but it suffers from a content leakage problem when combined with text conditions.
InstantStyle~\cite{instantstyle} suppresses content information in the CLIP embedding space and injects only the remaining style features, effectively removing all content cues.
Such complete removal of content is undesirable in fashion design, as users often expect the content of the image prompt (e.g., faces, object shapes, or other structural cues) to be reflected in the garment.

\noindent {\bf Generative AI-driven Fashion Design.}
Generative fashion image design aims to modify garment attributes (e.g., color, pattern, style) while preserving realism, wearer identity, posing significant challenges in balancing fidelity and realism. Text-driven methods like StyleCLIP~\cite{styleclip} utilize CLIP embeddings to alter generative model latents but often prioritize global text alignment, compromising local garment details. Broader diffusion-based approaches, such as instruction-tuned models~\cite{instructpix2pix} and mask-guided synthesis~\cite{paintbyexample}, offer general editing capabilities yet lack specialization in fashion contexts. Recent approaches like Text2Human~\cite{text2human} and TexFit~\cite{texfit} leverage textual prompts for garment specification but struggle with precise control over shape and style. Lots of fashion!~\cite{lots} combines a sketch (for outline) and text (for details) to accurately reflect a design concept, but it does not support VTON functionality.
Multimodal Garment Designer (MGD)~\cite{multimodalgarment} integrates sketch and text prompts for enhanced shape and color control, yet requires additional pose inputs, adding complexity and insufficiently capturing nuanced stylistic features like textures and materials. FashionTex~\cite{fashiontex} employs textual prompts for shape and fixed-size image patches for style, but it struggles with irregular and intricate garment details. PICTURE~\cite{picture} addresses this by allowing arbitrary garment image inputs for style extraction; however, it remains limited to garment-only design sources and clothing-only VTON. 

\begin{figure*}[h]
    \centerline{\includegraphics[width=1\textwidth]{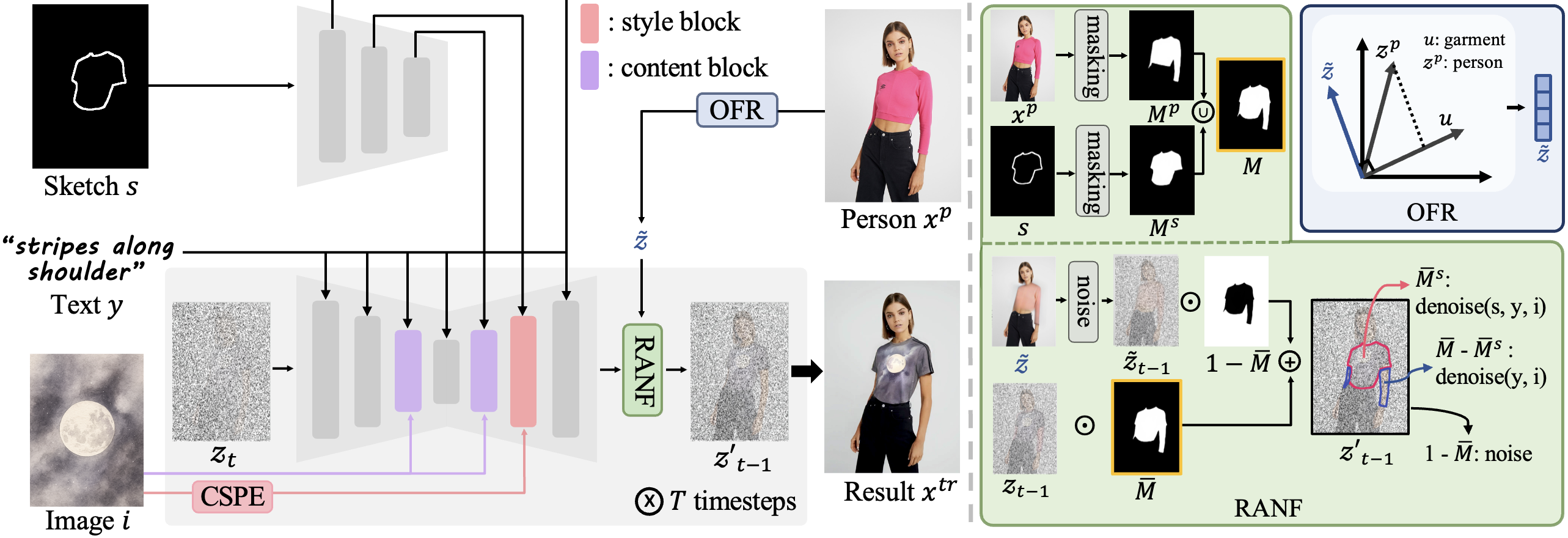}}
    \caption{Overview of our \textit{\textbf{FEAT} (\textbf{F}ashion \textbf{E}diting \textbf{A}nd \textbf{T}ry-On from Any Design).}} 
    \label{fig:pipeline}
\end{figure*}
\section{Method}
We introduce \textit{\textbf{FEAT} (\textbf{F}ashion \textbf{E}diting \textbf{A}nd \textbf{T}ry-On from Any Design)},
a novel method that
(1) enables diverse design not only from garments but also from non-apparel images such as artwork or natural photographs, and (2) supports holistic integration of  fashion items in a dataset-independent manner. To achieve these goals, we first propose a design method that separates content and style features in image prompts and selectively inject desired elements even from non-apparel images (Sec.~\ref{method:styling}). We further present a try-on method based on ControlNet~\cite{controlnet}, which removes regions of the original clothing that will be replaced through orthogonal projection and applies region-adaptive noise fusion tailored to the virtual try-on objective (Sec.~\ref{method:tryon}). 

Given a person image $x^{p}$, \textit{FEAT} generates a try-on result $x^{tr}$ by integrating a sketch $s$, an image prompt $i$, and a text prompt $y$. Our approach incorporates scaling factors to dynamically adjust the influence of each input modality, allowing for fine-grained control or complete exclusion of specific conditions—thereby enabling a more flexible, expressive, and realistic editing and try-on experience. An overview of the complete pipeline is shown in Fig.~\ref{fig:pipeline}. 

\subsection{Disentangled Dual Injection (DDI)}
\label{method:styling}
\noindent {\bf Selective Dual Injection (SDI).} 
Reducing the image scale in IP-Adapter can mitigate content leakage; however, because this parameter jointly modulates all image-derived signals, it consequently attenuates style information as well. To address this issue, we build upon the observation in InstantStyle~\cite{instantstyle} that individual UNet attention blocks exhibit distinct sensitivities to different attributes. Using fashion-domain data, we conduct a block-wise analysis of responsiveness and determine the block exhibiting the highest style sensitivity as the style block and the three blocks exhibiting the highest content sensitivity as the content blocks. In Fig.~\ref{fig:pipeline}, we present a simplified illustration of these designated blocks. Independent style and content scaling factors are then applied to these blocks, enabling precise and selective modulation of the corresponding components within the image prompt (Sec.~\ref{sec:experiment_ablation}).

\noindent {\bf Content-Subtractive Proxy Embedding (CSPE).} 
Block-level quantitative analysis (Sec.~\ref{sec:experiment_ablation}) reveals that InstantStyle’s style block still encodes substantial content information, indicating that assigning style or content scales alone is insufficient for achieving strict disentanglement. Moreover, InstantStyle subtracts content-related CLIP text embeddings from CLIP image embeddings, but imperfect image–text alignment often removes essential style cues. To address these limitations, we propose \textit{Content-Subtractive Proxy Embedding (CSPE)}, which suppresses content directly within the CLIP image embedding space. 
From an image prompt $i$, we derive a content proxy by preserving only its $L$ (lightness) channel and applying global blurring to eliminate color and texture. Subtracting the CLIP embedding of this proxy from the original embedding attenuates structural content while retaining stylistic attributes:
\begin{equation}
    \mathbf{e}_{\text{style}} = \phi(i) - \phi\!\left(\mathcal{B}_\sigma(\mathcal{L}(i))\right),
\end{equation}
where $\mathcal{L}(\cdot)$ extracts the $L$ (lightness) channel, $\mathcal{B}_\sigma$ denotes global blurring with standard deviation $\sigma$, and $\phi$ represents the CLIP image encoder. We inject $\mathbf{e}_{\text{style}}$ only into the style block, while the original embedding $\phi(i)$ is supplied to the content blocks. This design enables clear separation of the two attributes, allowing selective incorporation of desired elements into the final design.

\begin{table*}[t]
\centering
\renewcommand{\arraystretch}{1.25} 
\caption{Quantitative comparisons on garment and accessory datasets across various multimodal settings.}
\label{tab:main}
\resizebox{\textwidth}{!}{%
\begin{tabular}{l rrrrr rrrrr}
\hline
     & \multicolumn{5}{c}{\rule{0pt}{2.5ex} Garment Dataset}  
     & \multicolumn{5}{c}{Accessory Dataset} \\
\cmidrule(lr){2-6} \cmidrule(lr){7-11} 
     & \shortstack{GPT-4V $\uparrow$} & \shortstack{Sketch $\downarrow$} & \shortstack{Image $\uparrow$} & \shortstack{Text $\uparrow$} & \shortstack{Human $\uparrow$} & \shortstack{GPT-4V $\uparrow$} & \shortstack{Sketch $\downarrow$} & \shortstack{Image $\uparrow$} & \shortstack{Text $\uparrow$} & \shortstack{Human $\uparrow$} \\ \hline

\multicolumn{11}{l}{\textit{Sketch + Image}} \\ 
ControlNet+IP-Adapter & 1037.39 & 10.42 & 0.33 & - & 19.23\% & 1022.39 & 36.83 & 0.21 & - & 16.67\% \\
PICTURE  & 883.80 & 14.54 & 0.31 & - & 3.85\% & 853.80 & 125.37 & 0.19 & - & 8.33\% \\
\textbf{FEAT (ours)} & \textbf{1172.73} & \textbf{6.95} & \textbf{0.37} & - & \textbf{76.92\%} & \textbf{1207.73} & \textbf{8.30} & \textbf{0.31} & - & \textbf{75.00\%} \\ \hline

\multicolumn{11}{l}{\textit{Sketch + Text}} \\ 
ControlNet & 902.70 & 8.09 & - & 27.81 & 7.10\% & 1149.35 & 6.14 & - & 23.58 & 13.05\% \\
MGD  & 1027.48 & 7.32 & - & 28.35 & 18.43\%  & 814.02 & 38.31 & - & 22.65 & 6.42\% \\
\textbf{FEAT (ours)} & \textbf{1148.15} & \textbf{5.16} & - & \textbf{29.12} & \textbf{74.47\%} & \textbf{1210.31} & \textbf{4.00} & - & \textbf{25.18} & \textbf{80.52\%} \\ \hline

\multicolumn{11}{l}{\textit{Sketch + Image + Text}} \\ 
ControlNet+IP-Adapter & 912.13 & 9.04 & 0.30 & 27.48 & 19.23\% & 919.39 & 13.25 & 0.17 & 24.91 & 16.67\% \\
\textbf{FEAT (ours)} & \textbf{1087.86} & \textbf{4.83} & \textbf{0.36} & \textbf{28.40} & \textbf{80.77\%} & \textbf{1080.60} & \textbf{3.86} & \textbf{0.21} & \textbf{25.36} & \textbf{83.33\%} \\ \hline
\end{tabular}
}
\end{table*}

\subsection{Orthogonal-Guided Noise Fusion (OGNF)}
\label{method:tryon}
VTON differs fundamentally from inpainting for missing-region restoration, as it must simultaneously (i) preserve the person's pose and facial identity, (ii) remove the garments to be replaced, and (iii) synthesize new garments in a coherent manner~\cite{vton, kim2025diverse}. To meet these objectives jointly, we introduce \textit{Orthogonal Fashion Removal (OFR)} and \textit{Region-Adaptive Noise Fusion (RANF)}.

\noindent {\bf Orthogonal Fashion Removal (OFR).}  
For clarity, we describe \textit{OFR} using garments as the reference item, although the same procedure applies to other fashion items. \textit{OFR} selectively removes garment information from the latent representation of the input person image without impairing non-garment regions such as the face, body, and background. Given an input person image $x^{p}$ and its garment segmentation mask $g$, we encode them using the VAE encoder to obtain $z^{p}$ and $z^{g}$, respectively. We additionally encode a white image $w$ to obtain $z^{w}$. Because the segmented garment $g$ contains a white background, subtracting $z^{w}$ from $z^{g}$ eliminates background contributions and isolates the latent direction corresponding to garment characteristics. We define this garment-specific latent direction as follows:
\begin{equation}
    v = z^{g} - z^{w}.
\end{equation}
We normalize this direction as $u = \frac{v}{\|v\|}$ and suppress garment-related components in $z^{p}$ by subtracting its projection onto $u$, resulting in the garment-attenuated latent representation $\tilde{z}$:
\begin{equation}
    \tilde{z} = z^{p} - \alpha \,(z^{p} \cdot u)\, u,
\end{equation}
where $\alpha$ controls the removal strength. This operation allows \textit{OFR} to produce a latent representation in which the target garment is largely eliminated, providing a stable foundation for subsequent synthesis.

\noindent \textbf{Region-Adaptive Noise Fusion (RANF).}
Using the garment-suppressed latent representation $\tilde{z}$ obtained through \textit{OFR}, we address the three core requirements of VTON—identity preservation, complete removal of existing garments, and synthesis of new garments—by dividing the input into three regions and applying region-specific noise strategies. From the input person image $x^{p}$, we extract the garment mask $M^{p}$, and from the sketch $s$, we obtain the sketch mask $M^{s}$. The overall manipulation region is defined as:
\begin{equation}
    M \;=\; M^{p}\,\cup\,M^{s},
\end{equation}
where all masks are downsampled to the VAE-encoder resolution to yield latent masks $\bar{M}^{p}, \bar{M}^{s}, \bar{M}$. Across $T$ inference steps, the latent state $z'_{t-1}$ is computed by combining regions according to the following scheme:
\begin{equation}
    \label{eq:ranf_update}
    \begin{aligned}
    z'_{t-1}
    &= \bar{M}\odot \mathrm{denoise}\!\left(z_t, s, y, i, t\right) \\
    &\quad + \left(\mathbf{1}-\bar{M}\right)\odot \mathrm{noise}\!\left(\tilde{z} , t\right).
    \end{aligned}
\end{equation}

The resulting $z'_{t-1}$ is then fed into the next denoising step, and iterating this process for $T$ steps yields distinct behaviors across regions:
\begin{itemize}
    \item \textbf{$\bar{M}^{s}$ (New Garment Synthesis Region):} New garments are synthesized using the sketch $s$, ensuring high fidelity to the specified shape.
    \item \textbf{$\bar{M} - \bar{M}^{s}$ (Existing Garment Removal Region):} Since sketch guidance is applied only within $M^{s}$, this region is denoised without sketch-based guidance, effectively removing the original garment.
    \item \textbf{$1 - \bar{M}$ (Non-Garment Region):} Inspired by Repaint~\cite{repaint}, noise re-injection preserves visual consistency in the face, body, and background.
\end{itemize}

This region-adaptive formulation introduces, to the best of our knowledge, the first three-region design that employs distinct noise-control strategies in each region, tailored to VTON requirements. This design enables realistic try-on results without pairwise training data. Beyond human photographs, the proposed framework also transfers effectively to non-human visual domains—such as game characters and animated figures—demonstrating its robustness to variations in rendering design and its broader applicability across heterogeneous visual modalities, as shown in Fig.~\ref{fig:domain}.

\section{Experiments}
\subsection{Implementation and Evaluation Details}
\label{sec:experiment_setup}

\noindent {\bf Dataset.} 
For evaluation, we collect garment sketch inputs from two virtual try-on datasets: VITON-HD~\cite{vitondataset}, which focuses on tops, and DressCode~\cite{dresscodedataset}, which includes tops, bottoms, and dresses. We extract 300 sketches from VITON-HD and 600 from DressCode. Additionally, we collect 100 accessory sketches—25 each for bags, shoes, scarves, and belts—from DressCode, resulting in a total of 1,000 sketches. We generate 1,000 text prompts using GPT-4V and randomly collect 1,000 image prompts from WikiArt~\cite{wikiart} (200 samples) and Midjourney (800 samples). Ablation studies are conducted using data drawn from DressCode, as it covers a broader range of garment types. Further details on dataset construction are provided in the supplementary material.

\begin{figure*}[t]
  \centerline{\includegraphics[width=\textwidth]{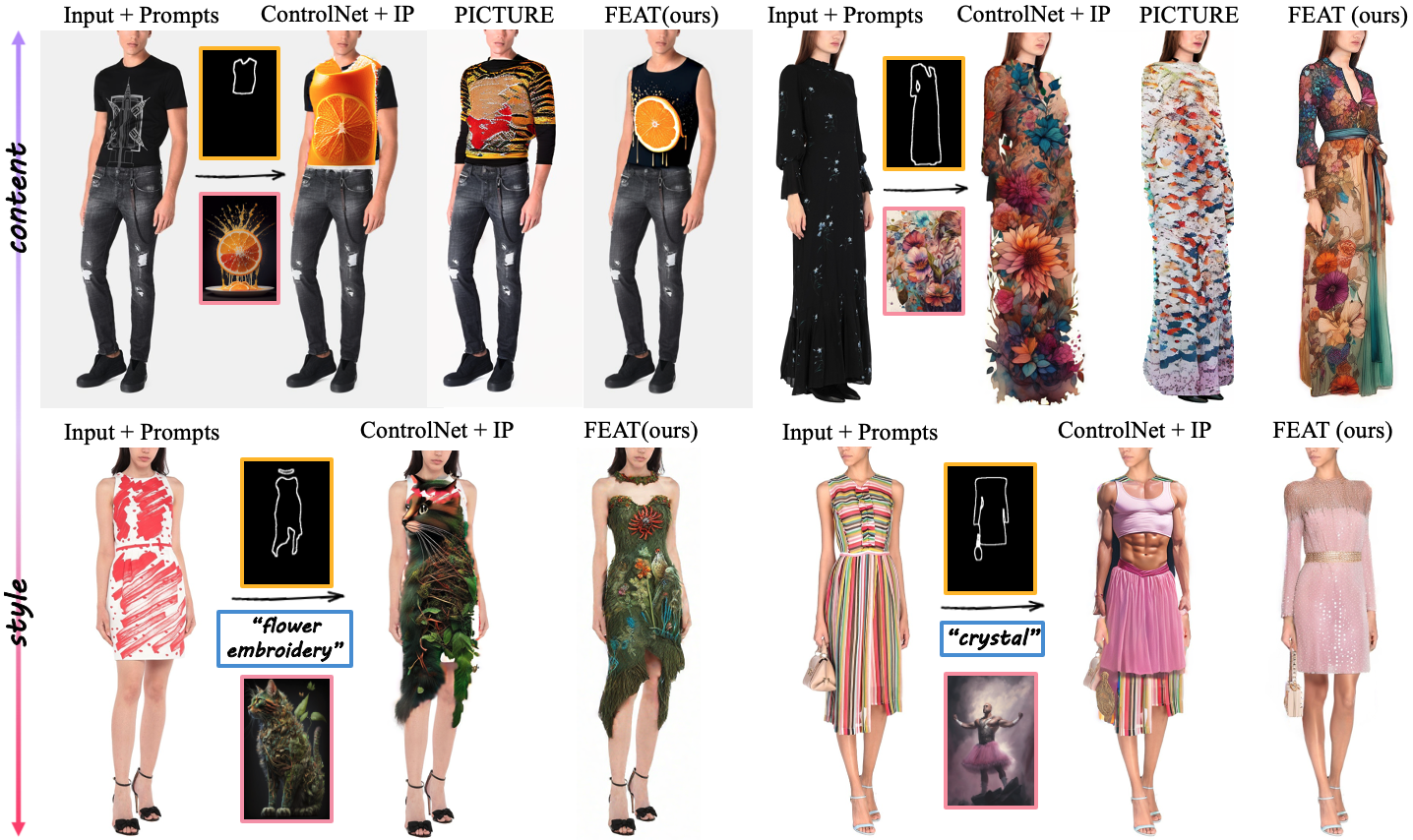}}
    \caption{Qualitative comparisons under content and style conditioned settings.}
    \label{fig:qualitative}
\end{figure*}

\noindent {\bf Baselines.}
We compare our method against four alternatives adapted for multimodal fashion control. The first baseline combines ControlNet~\cite{controlnet} and IP-Adapter~\cite{ip-adapter}, where sketches and text prompts are processed by ControlNet, while image prompts are handled by IP-Adapter to enable joint conditioning. The second baseline is PICTURE~\cite{picture}, a virtual try-on model that accepts structural and visual features as input. For fair comparison, we convert sketches into garment images using IP-Adapter and use the resulting images as the structural condition for PICTURE. The third is Multimodal Garment Designer (MGD)~\cite{multimodalgarment}, which supports fashion editing from sketch and text prompts. Lastly, we include ControlNet~\cite{controlnet} integrated into Stable Diffusion XL~\cite{sdxl} for image manipulation guided by sketch and text inputs.

\noindent {\bf Implementation.}
All experiments were conducted using a single NVIDIA A6000 GPU. For fair comparison, all models were evaluated using the DDIM sampler with 50 inference steps and a fixed classifier-free guidance scale of 7.5. For control signal extraction, we employed ControlNet-Depth trained on Stable Diffusion XL, and the IP-Adapter used for image prompts was also based on the Stable Diffusion XL. For multimodal conditioning, the scales for sketch, image, and text prompts were set to 0.7, 0.5, and 0.5, respectively.

\noindent {\bf Metrics.}
To assess overall quality, we adopt the Elo score evaluation framework proposed in GPTEval3D~\cite{gpteval3d}, using GPT-4V as the evaluator. Unlike conventional metrics that focus on a single aspect (e.g., text similarity), this approach enables a more holistic and human-aligned assessment of multimodal consistency and visual realism. We extend this method to the VTON setting. We additionally report Chamfer Distance (CD)~\cite{sketch2human}, style score~\cite{less_is_more}, and CLIP-T~\cite{clip} to quantify how well each modality is reflected in the output. We also conduct a user study~\cite{lee2022bridging} to evaluate human preference.

\begin{itemize}
    \item GPT-4V (Elo): We adopt the Elo evaluation protocol from GPTEval3D~\cite{gpteval3d}, which combines meta-prompting with pairwise image comparisons to assess relative output quality. The meta-prompt framework dynamically generates evaluation prompts to capture diverse criteria, enhancing both accuracy and consistency. GPT-4V compares each image pair and assigns Elo scores based on relative preference. The specific prompts used are provided in the supplementary material. Evaluation is based on three criteria: (1) faithful reflection of all multimodal conditions, (2) preservation of the input identity, and (3) overall visual realism.

    \item Sketch (CD): To assess sketch alignment, we compute the CD between the input sketch and an edge map extracted from the generated garment region. 

    \item Image (style score): To quantify how faithfully the generated image follows the input image prompt without content leakage, we adopt the style score proposed in~\cite{less_is_more}. A lower score indicates greater content leakage, while a higher score reflects better similarity between the generated result and the image prompt.

    \item Text (CLIP-T): CLIP text score is used to measure textual consistency between the generated image and the input text prompt.

    \item Human (Preference): We conduct a user study with 21 participants, asking them to select the image with the highest overall multimodal consistency and visual realism.
\end{itemize}

\subsection{Performance Evaluation}
\label{sec:experiment_quantitative }
\noindent {\bf Quantitative Analysis.} 
Tab.~\ref{tab:main} presents quantitative comparisons using content and style scales of 0.5, where the image prompt retains both content and style information. Results using only style information (excluding content) are provided in the supplementary material. \textit{FEAT} outperforms all baselines on both garment and accessory datasets, demonstrating its overall effectiveness. In particular, the large improvement in the Sketch score indicates that the original clothing is effectively removed while the sketch’s shape cues are faithfully reflected. Moreover, strong performance in both Image and Text scores confirms that content leakage is well suppressed, achieving balanced integration across all conditioning modalities.

\begin{figure}[t]
  \centerline{\includegraphics[width=0.5\textwidth]{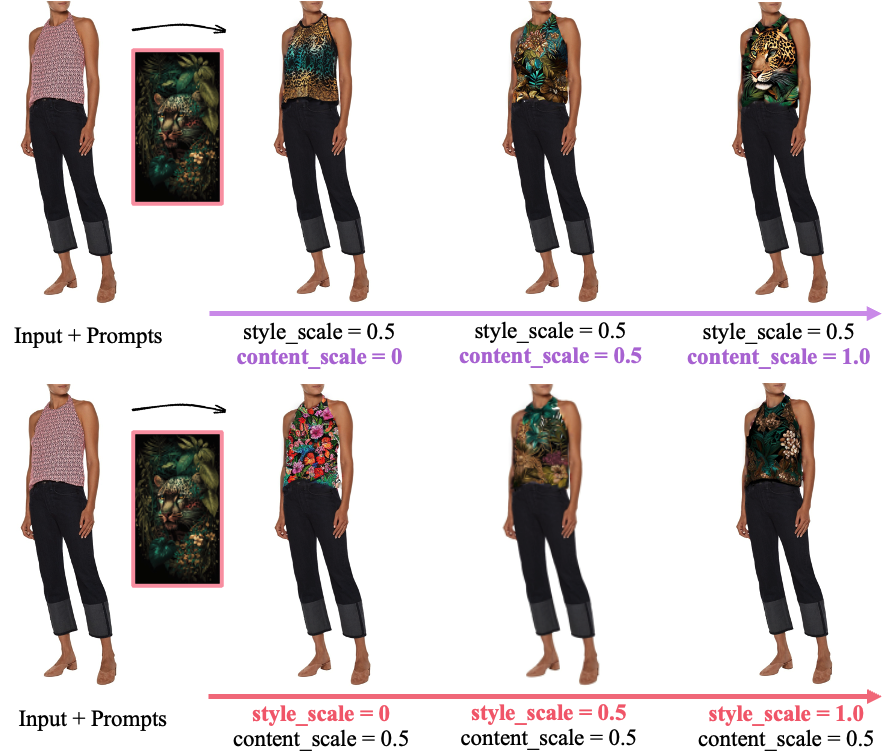}}
    \caption{Visual comparisons of scaling factor variations.}
    \label{fig:ablation_scale}
\end{figure}

\begin{figure}[t]
\centering
\includegraphics[width=0.46\textwidth]{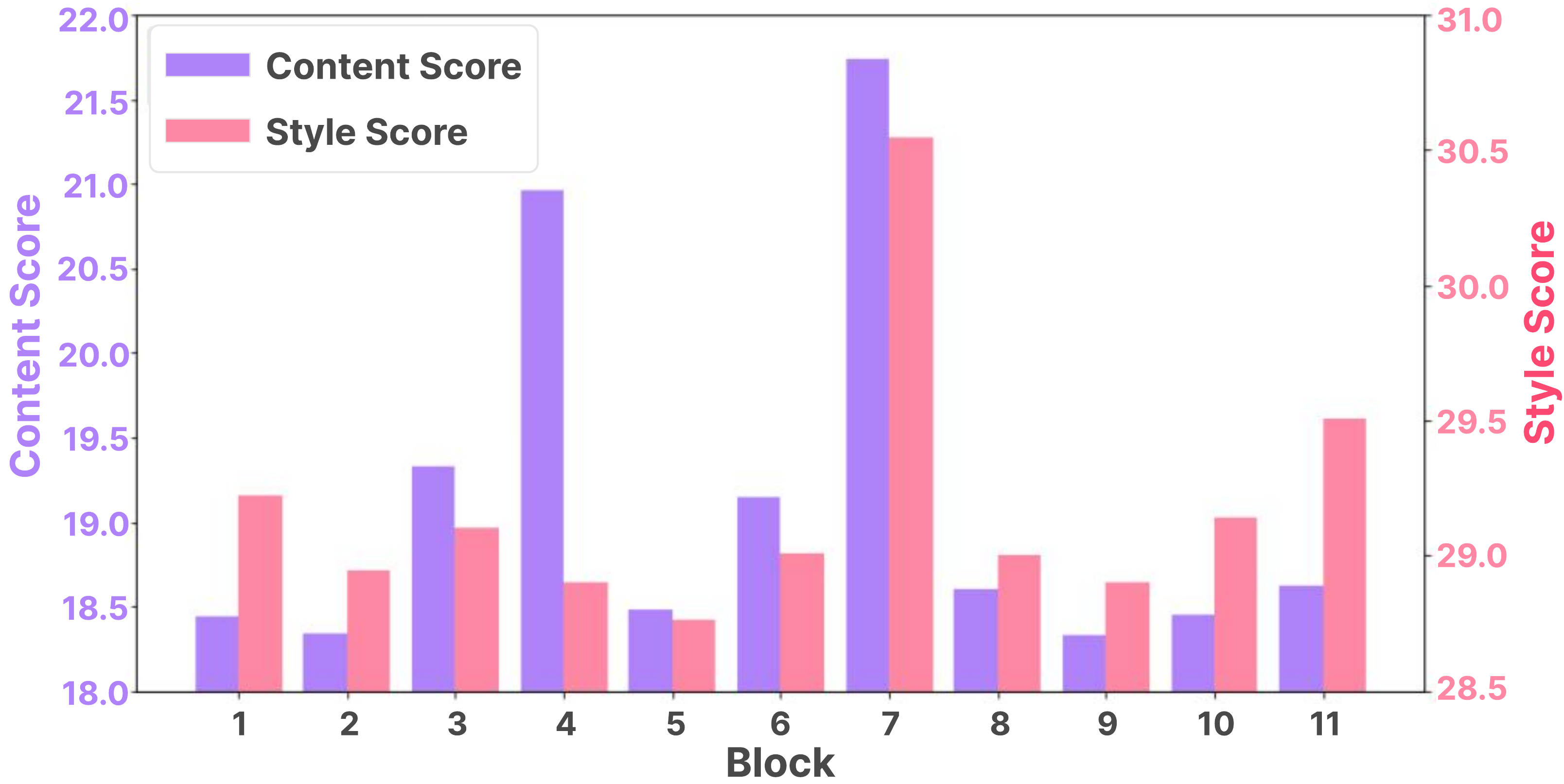}
\caption{Ablation study on block
injecting the IP-Adapter~\cite{ip-adapter} feature into individual attention blocks.} 
\label{fig:ablation_blocks}
\end{figure}

\noindent {\bf Qualitative Analysis.} 
\label{sec:experiment_quanitative}
The first row of Fig.~\ref{fig:qualitative} shows qualitative comparisons using both content and style from the image prompt. \textit{FEAT} generates natural and realistic try-on results that faithfully reflect the sketch and image prompt. In contrast, ControlNet + IP-Adapter leaves garment residues and suffers from strong content leakage, weakening sketch guidance and producing unnatural outputs. PICTURE avoids content leakage but fails to capture the visual information of the image prompt and cannot handle accessories. The second row of Fig.~\ref{fig:qualitative} presents results using style-only conditioning. \textit{FEAT} achieves balanced fusion across sketch, image, and text prompts, enabling coherent garment and accessory replacement. Meanwhile, ControlNet + IP-Adapter not only leaves residual garment and bag but also transfers unintended body-shape cues—a form of content leakage—such as the muscular male torso in the second-row rightmost example, which distorts the guidance from other modalities.

\begin{figure}[t]
  \centerline{\includegraphics[width=0.5\textwidth]{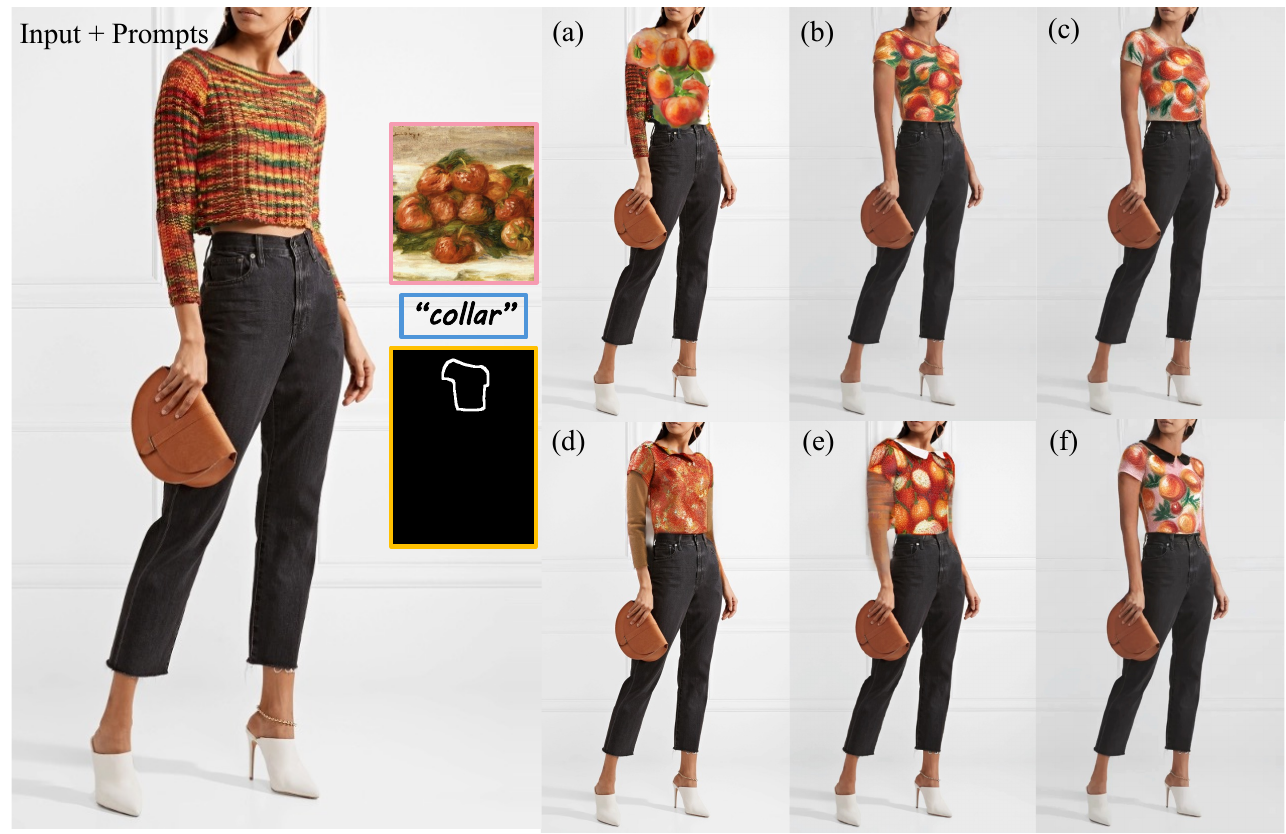}}
    \caption{Visual comparisons of ablation study.} 
    \label{fig:ablation_method}
\end{figure}

\begin{table}[t]
\centering
\setlength{\tabcolsep}{2.5pt}
\renewcommand{\arraystretch}{1.05}
\caption{Quantitative comparisons of ablation study on \textit{FEAT}.}
\label{tab:ablation_method}

\begin{tabularx}{0.48\textwidth}{@{}l l *{3}{>{\centering\arraybackslash}X}@{}}
\toprule
& Component  & \makecell{Sketch $\downarrow$} 
       & \makecell{Image $\uparrow$} 
       & \makecell{Text $\uparrow$} \\
\midrule
Baseline
  & (a) ControlNet + IP   & 9.71 & 0.28 & 27.01 \\
\midrule
\multirow{2}{*}{\textit{DDI}}
  & (b) w/o \textit{SDI}           & 4.89 & 0.30 & 27.58 \\
  & (c) w/o \textit{CSPE}          & 4.52 & 0.29 & 27.42 \\
\midrule
\multirow{2}{*}{\textit{OGNF}}
  & (d) w/o \textit{OFR}           & 5.57 & 0.34 & 27.79 \\
  & (e) w/o \textit{RANF}          & 5.28 & 0.33 & 27.72 \\
\midrule
Full
  & (f) \textbf{\textit{FEAT}}     & \textbf{4.15} & \textbf{0.36} & \textbf{27.88} \\
\bottomrule
\end{tabularx}
\end{table}

\subsection{Ablation Study}
\label{sec:experiment_ablation}
\noindent {\bf The Effect of Varying Content \& Style Scale.}
Fig.~\ref{fig:ablation_scale} presents qualitative results illustrating the effect of content and style scales. In the first row, we fix the style scale to 0.5 and increase the content scale. When the content scale is 0, no content information is reflected; at 0.5, small content elements such as leaves and flowers begin to appear; and at 1.0, large content cues like the tiger are clearly expressed. In the second row, we fix the content scale to 0.5 and increase the style scale, observing progressively stronger stylistic characteristics. These results demonstrate that our \textit{DDI} effectively disentangles content and style, allowing users to selectively control each component.

\noindent {\bf The Effect of Content \& Style Blocks.}
The attention layers in the SDXL U-Net are grouped into 11 blocks. As shown in Fig.~\ref{fig:ablation_blocks}, we inject image-prompt features into each block using IP-Adapter~\cite{ip-adapter} with the image scale set to 1.0, and evaluate content and style scores. The content score is computed using CLIP-Text similarity with GPT-4V–extracted content words, while the style score follows StyleScore~\cite{dreamstyler}. The results show that Block 7—defined as the style block in InstantStyle—obtains the highest score not only in style but also in content, indicating that it mixes strong style and content signals. Block 4 shows the second-highest content score while having a notably low style score. Blocks 3 and 6 also exhibit relatively high content scores.

\noindent {\bf The Effect of \textit{FEAT} Components.}
Fig.~\ref{fig:ablation_method} and Tab.~\ref{tab:ablation_method} present qualitative and quantitative results for an ablation study of \textit{FEAT}. 

Specifically, (a) shows the result of performing virtual try-on using only ControlNet and IP-Adapter, without any of our proposed components. The content from the image prompt is overly reflected, causing visual distortions that are not perceived as clothing, and the original garment remains visible, indicating a failure to properly replace it.
In (b), after removing \textit{SDI}, the result shows slightly less content leakage than (a), but the image prompt content is still too dominant, leading to insufficient influence from the text prompt. (c) shows the result without \textit{CSPE}. This outcome is similar to (b), with an overemphasis on the image content and an imbalance where the text prompt’s effect is underrepresented. 
The comparison between (b) and (c) indicates that both \textit{SDI} and \textit{CSPE} are necessary for the three input signals (sketch, image, and text) to be integrated harmoniously without interfering with each other. 

In (d), without \textit{OFR}, the original clothing is not removed and remains blended with the newly applied garment. 
(e) shows the result without \textit{RANF}. In this case, the original outfit is still not completely eliminated, demonstrating that the combination of \textit{OFR} and \textit{RANF} is essential for fully removing the original garment. 
Consequently, the full \textit{FEAT} (f), which includes all proposed components, successfully replaces the original clothing with the new attire, achieving the highest design quality and VTON performance with all input modalities harmoniously integrated.

\begin{figure}[t]
  \centerline{\includegraphics[width=0.46\textwidth]{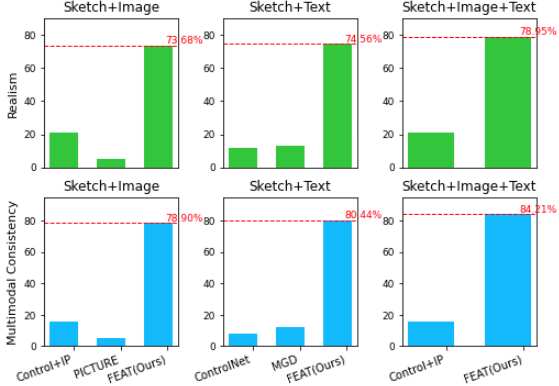}}
    \caption{User study results across different multimodal settings.}
    \label{fig:userstudy}
\end{figure}

\subsection{User Study}
To evaluate the perceptual quality of our proposed method, we conducted a user study with 21 participants as seen in Fig~\ref{fig:userstudy}. A total of 57 examples were presented, each containing results from different models. For each example, participants were asked to select the most appropriate image according to one of the following criteria. 
(i) Realism: How natural and realistic the generated image appears as a virtual try-on result. (ii) Multimodal Consistency: How well the generated image aligns with the given multimodal inputs (sketch, image, and text).

The study was conducted under three different input settings: Sketch+Image+Text, Sketch+Text, and Sketch+Image. In all settings, our method achieved the highest selection rate, demonstrating superior perceptual quality compared to existing baselines. While \textit{FEAT} performed well in terms of multimodal coherence, its improvements in realism were even more pronounced.
The study was conducted under the setting where image prompts retained both content and style cues. Additional results under a style-only setting are provided in the supplementary material.

\begin{figure}[t]
  \centerline{\includegraphics[width=0.55\textwidth]{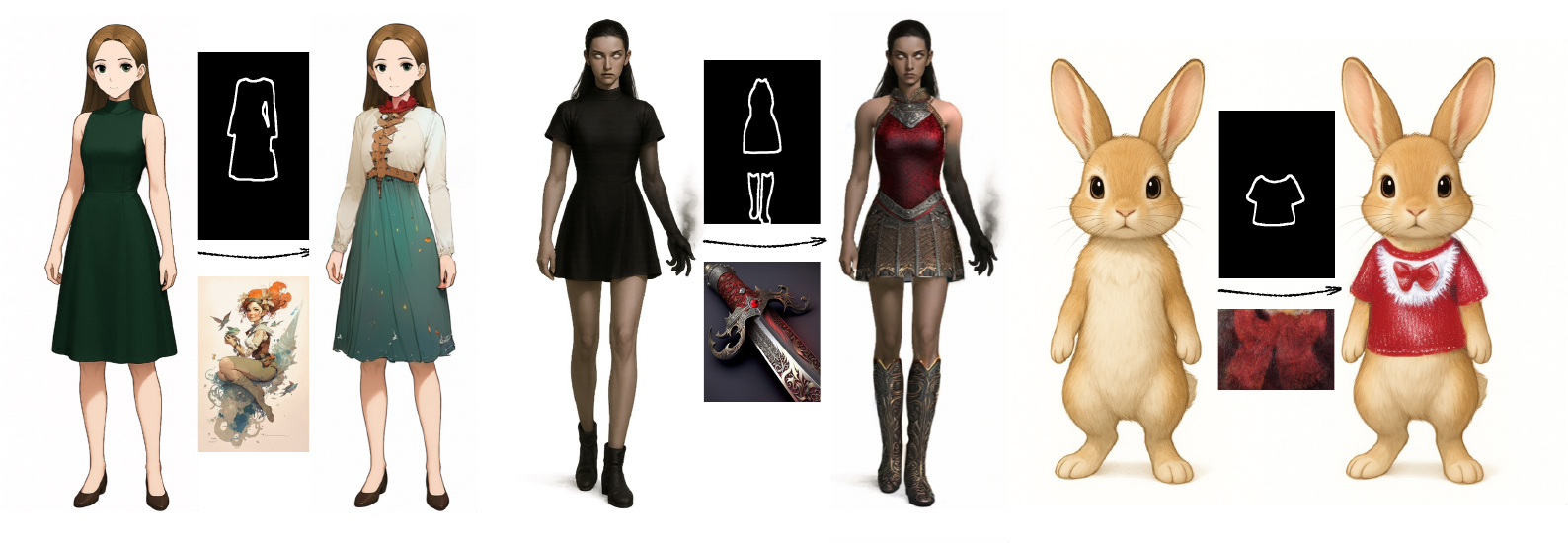}}
    \caption{Visualization of cross-domain generalization for editing and try-on using \textit{FEAT}.} 
    \label{fig:domain}
\end{figure}

\subsection{Cross-Domain Applicability}
\label{sec:additional_results}
Owing to its training-free design, \textit{FEAT} can be applied not only to the conventional human-photo-based fashion editing domain (commonly used for fashion editing and virtual try-on) but also to a wide variety of other design domains. As shown in Fig.~\ref{fig:domain}, our method performs fashion editing and VTON in a visually coherent, artifact-free manner across diverse domains, including animation, game characters, and even animals. This demonstrates the high practical versatility of our approach and suggests that it can be readily applied to various scenarios.

\section{Conclusion}
We presented \textit{FEAT}, a novel approach for comprehensive fashion editing and VTON that supports diverse design sources and full outfits, including garments and accessories. Our \textit{DDI} module enables selective integration of content and style from image prompts, while \textit{OGNF} removes residual garments via orthogonal projection and applies region-aware noise processing to produce natural try-on results without pairwise dataset training. Extensive experiments across various settings demonstrate that \textit{FEAT} significantly outperforms existing baselines and achieves state-of-the-art performance in multimodal prompt consistency and visual realism. A remaining limitation is that rendering very small accessories near the face can be unstable; we believe this can be mitigated in future work by incorporating localized refinement modules.

\section*{Acknowledgment}
This work was supported by the National Research Foundation of Korea (NRF) grant (RS-2025-00555943); by the AI Computing Infrastructure Enhancement (GPU Rental Support) User Support Program (RQT-25-090040); and by the Institute of Information \& Communications Technology Planning \& Evaluation (IITP) grants (No.RS-2025-02219317; AI Star Fellowship (Kookmin University), IITP-2024-RS-2024-00397085; Leading Generative AI Human Resources Development, and IITP-2024-RS-2024-00417958; Global Research Support Program in the Digital Field) funded by the Korea government (MSIT).

{
    \small
    \bibliographystyle{ieeenat_fullname}
    \bibliography{main}
}


\end{document}